\documentclass{article}

\usepackage[final]{neurips_2023}

\usepackage[utf8]{inputenc} 
\usepackage[T1]{fontenc}    
\usepackage{hyperref}       
\usepackage{url}            
\usepackage{booktabs}       
\usepackage{amsfonts}       
\usepackage{nicefrac}       
\usepackage{microtype}      
\usepackage{xcolor}         
\usepackage{graphicx}
\usepackage{amsmath}
\usepackage{multirow}
\usepackage{comment}
\usepackage{float}

\title{CarbonNet: How Computer Vision Plays a Role in Climate Change? \\ Application: Learning Geomechanics from Subsurface Geometry of CCS to Mitigate Global Warming}

\author{%
  Wei Chen\thanks{Corresponding author} \\
  Stanford University \\
  \texttt{weichen6@stanford.edu} \\
  \And
  Yunan LI \\
  Stanford University \\
  \texttt{yunanli@stanford.edu} \\
  \AND
  Yuan TIAN \\
  Stanford University \\
  \texttt{ytian96@stanford.edu} \\
}

\begin{document}

\maketitle

\begin{abstract}
    We introduce a new approach that uses computer vision to predict land surface displacement from subsurface geometry images for carbon capture and sequestration (CCS). CCS has been proven to be a key component of a carbon-neutral society. However, scientists face several challenges along the way, including the high computational cost caused by the large model scale and the limited ability of pre-trained models to generalize to complex physics. We tackle these challenges by training models directly on subsurface geometry images. The goal is to understand the response of land surface displacement to carbon injection and to use our trained models to inform decision making in CCS projects.

  We implement multiple models (CNN, ResNet, and ResNetUNet) for the static mechanics problem, which is an image prediction problem. We then use LSTM and transformer models for the transient mechanics scenario, which is a video prediction problem. The results show that ResNetUNet outperforms the other models on the static mechanics problem, thanks to its architecture, and that the LSTM shows performance comparable to the transformer on the transient problem. This report proceeds by describing our dataset in detail, followed by the model descriptions in the Methods section. The Results and Discussion sections state the key learnings and observations, and the Conclusion with future work rounds out the paper.
\end{abstract}

\section{Introduction}

\begin{figure*}[h]
    \centering
    \includegraphics[width = 0.8\textwidth]{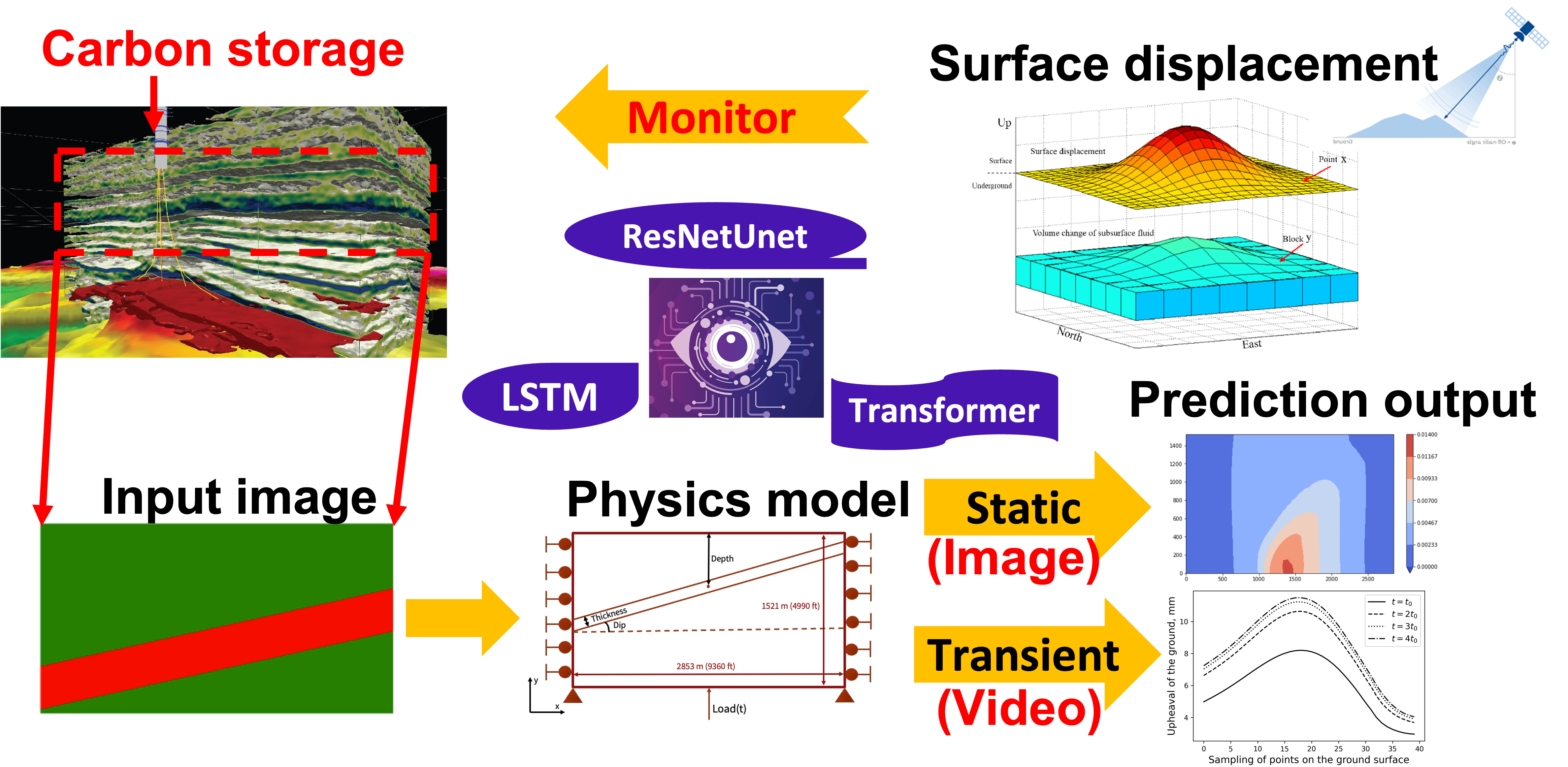}
    \caption{A summary of the workflow of this project, whose goal is to leverage computer vision intelligence for carbon storage surveillance in order to mitigate climate change. ResNetUNet assists the image prediction of land surface displacement, and we use LSTM and transformer models to predict displacement changes over time, which is a video problem in the computer vision domain.}
    \label{fig:workflow}
\end{figure*}

CCS is an important emissions-reduction technology for alleviating global warming and climate change \cite{stephens2006growing}. In CCS, supercritical $\mathrm{CO_2}$ is injected into deep brine aquifers or abandoned oil reservoirs for permanent storage. However, a possible risk is surface upheaval induced by excessive fluid pressure, which may damage structures on the ground and affect people's lives \cite{li2013application}. Understanding and predicting the spatial distribution and temporal change of surface deformation is important for avoiding such hazards. However, this is challenged by the large spatial scale and long time scale of the project, which make direct numerical simulation computationally prohibitive. Because traditional numerical simulation usually requires
extensive computing resources \cite{Vilarrasa2010, Shi2013, Li2016, Fuchs2019, Talebian2013}, neural networks may provide a less accurate but more efficient and sufficient solution that benefits model-based decision making and planning \cite{Tang2021}. We formulate the entire project as the workflow described in Fig. \ref{fig:workflow}. The objective of the project is to predict the variation of surface vertical displacement (upheaval) with time induced by the injected fluid under different subsurface geometry settings. The inputs are 2D images of the subsurface geometry, consisting of a shale layer embedded at different dip angles between two permeable layers. The output is the displacement image. The shale layer and the permeable layers exhibit a large contrast in their mechanical property, the stiffness $\mathbb{C}^e$, and their flow property, the permeability $\boldsymbol{\kappa}$ \cite{josh2012laboratory}.

In this project, we are motivated to apply deep learning models to predict the land surface displacement under different subsurface geometries in a coupled geomechanics and flow problem. We start from the static mechanics model, where the fluid pressure is treated as a prescribed traction boundary condition. For prediction with the static mechanics model, the input to our algorithm is a geometry image. We then use different models (CNN, ResNet, ResNetUNet) to output a predicted displacement image (Fig. \ref{fig:workflow}, static prediction output) across the domain of the input geometry image. ResNetUNet is designed by us to combine the advantages of the UNet and ResNet architectures to map different subsurface geometries to the distribution of the 2D displacement field.

From the ResNetUNet in the static mechanics model, we proceed to the LSTM and transformer models to predict the temporal change of the surface displacement. The input to our algorithm is a geometry image along with five frames of displacement curves, which is in the format of a short video. We then use the LSTM and transformer to output the predicted displacement curve (Fig. \ref{fig:workflow}, transient prediction output) at the next time step. LSTM is a typical algorithm for time series predictions, and we are motivated to test the transformer because of its self-attention mechanism. It offers the possibility of predicting long-term dependencies by contracting the distance between any two locations to a constant, thereby overcoming the issue of information loss in the sequential computation of the LSTM \cite{zeyer2019comparison}.

\section{Related Work} 
Deep learning approaches have been proven effective in predicting the static and time-dependent dynamics of a system governed by physics principles expressed as partial differential equations (PDEs). The existing deep learning approaches are of two types: physics-informed neural networks (PINN) \cite{Xu2020, Meng2020, Zheng2020} and simulation-data-driven approaches \cite{Tang2021, Tang2020, Wen2021,nie2020stress}. The difference lies in whether direct simulation data are needed to train the model.

The PINN method obtains an approximate solution by minimizing the residual of the governing physics equations. In contrast, common numerical discretization methods use the finite difference method (FDM) and the finite element method (FEM). PINN is a mesh-free method that employs automatic differentiation to handle differential operators. The existing PINN model for the coupled flow and geomechanics problem \cite{bekele2020deep} is limited to one-dimensional problems, and it ignores the influence of material heterogeneity. Besides, it only predicts the excess pore pressure on the flow side, but not the surface displacement on the geomechanics side. The other type of deep learning approach is the simulation-data-driven method, which takes the solution from simulation directly as training data. The end-to-end image-based approaches take figures with the model geometry as input and produce contour plots showing the distribution of different physical quantities (fluid pressure, displacement) as output. A ResNet CNN was applied to predict the stress field of a 2D cantilever beam under different boundary conditions \cite{nie2020stress}. Nevertheless, it is limited to static solid mechanics problems and ignores the coupling between fluid flow and the mechanical response. Besides, it only deals with homogeneous models, which is not the case in reality. The R-U-Net model with LSTM has been used to predict the temporal and spatial distributions of surface displacement \cite{Tang2021}, fluid pressure, and saturation in 3D models. However, it is restricted to predicting one specific geological formation setting. 

In order to predict the temporal change of the surface displacement under different geological formation settings, a more advanced model is needed to capture the dynamics changing with time. Transformer and self-attention models \cite{NIPS2017_3f5ee243}, which were originally developed in NLP, have shown the ability to outperform other methods by learning longer-term dependencies without recurrent connections. The transformer has been used to predict representative dynamical systems such as 2D fluid dynamics and 3D reaction-diffusion dynamics \cite{nie2020stress}. The model shows its effectiveness in capturing the varying behavior of the evolution of different physical phenomena. Therefore, the transformer is also chosen in our project to predict the transient behavior of the surface displacement under different subsurface geometry settings.

\section{Dataset}

\begin{figure}[h]
    \centering
    \includegraphics[width = 0.7 \textwidth]{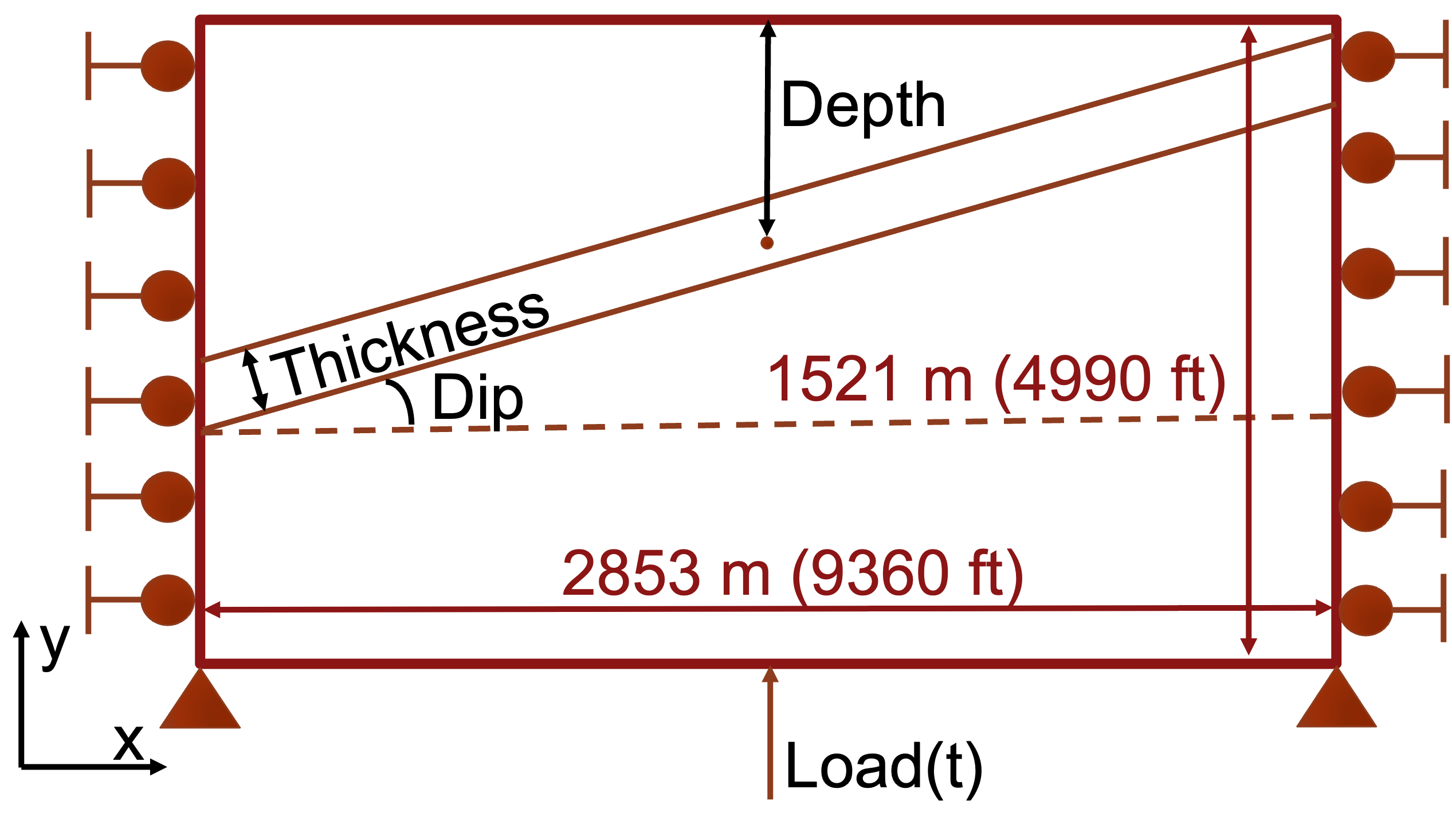}
    \caption{The layout of the computational problem. The purpose of the computational program is to capture the displacement given the configuration of the loading, the material distribution, and the boundary conditions \cite{bathe2007finite}. }
    \label{fig:fem_layout}
\end{figure}

In this section, we describe the dataset used in the project. The dataset includes two parts: the static model and the transient model. The input dataset consists of 2D geometries of geological layers (Fig. \ref{fig:workflow}, input image), which comprise two heterogeneous materials: hard rock with high permeability and soft shale with low permeability. The label is the surface displacement contour (Fig. \ref{fig:workflow}, static prediction output image) under a specific loading. The geometries are sampled according to real underground strata. The size of the domain (the length and width of the rectangle) and the thickness and depth of the shale layer are fixed, while the dip angle of the shale layer is varied. The surface displacement field data are generated from simulation results in \href{https://fenicsproject.org/}{FEniCS} (an open-source computational platform using the finite element method (FEM)) \cite{alnaes2015fenics}. The model setting used in the problem is depicted in Fig. \ref{fig:fem_layout}. Data augmentation is not necessary in this project because we can generate as many training samples as we want from the physics simulation.

\subsubsection{Geometry as the input}
The geometries are generated as the input of the machine learning model using the \texttt{PIL.draw} module and are depicted in Fig. \ref{fig:workflow}, labeled as the input image. The material characteristics of the shale and the rock are distinct; thus, different layouts of the materials yield different results. We generate 10,000 groups of data for training by sampling the dip angle uniformly from 0$^\circ$ to 45$^\circ$. Different dip angles correspond to specific geometries, which affect the resulting displacement. We split the entire dataset into 95\% for training and 5\% for validation. In addition, we generated 5\% of new samples for testing. 

\subsubsection{Static displacement as the label}
We consider the vertical displacement $u_y$ as the label; it is defined on each vertex of the finite element mesh \cite{ho1988finite} and can be considered as the displacement at points uniformly sampled over the domain. The vertical displacement ($u_y$) corresponds to the upheaval of the stratum. The FEniCS program provides the following API:
\begin{equation}
    \text{Layout of problem} \Longrightarrow \text{$u_y$ across the domain}
\end{equation}
As for the labels (ground truth), we use a \texttt{numpy} 1D array to store the vertical displacement of 1250 points from a uniform $50\times 25$ grid of points sampled over the domain. The dataset is summarized in Tab. \ref{tab:1}.

For the mathematical formulation, refer to Appendix A. 

\begin{table}[h]
\begin{center}
\caption{The dataset for the static machine learning model.}
\label{tab:1}
\begin{tabular}{  c | c} 
\hline
 input image & ground truth/label \\
 \hline
  Input image with dip angle $0^\circ$ & numpy array (1250 size)  \\ 
 \hline
  $\vdots$$\quad$ (9998 examples) & $\vdots$$\quad$ (9998 examples)\\ 
 \hline
  Input image with dip angle $45^\circ$ & numpy array (1250 size)\\ 
 \hline
\end{tabular}
\end{center}
\end{table}

\subsubsection{Transient displacement as the label}
Building on the static model, the next problem we address is the coupled hydro-mechanical simulation, which takes the fluid pressure into account to better represent the real situation. The coupled hydro-mechanical simulation is widely applied in industry and is well known for handling consolidation, which is a transient (time-dependent) problem. We use the same kind of geometry as the one generated for the static case, but employ another set of equations to simulate the physical process. The mathematical formulation can be found in Appendix A. 

In the static case, we compute a single displacement field across the domain given one input geometry. In contrast, the transient simulation yields multiple images at different time steps given one single geometry. In addition, we focus on the vertical displacement at the ground surface, which is the quantity of interest in industry. In this project, we sample 40 points on the ground surface and record their vertical displacement.

\begin{figure*}[h]
    \centering
    \includegraphics[width = 0.85\textwidth]{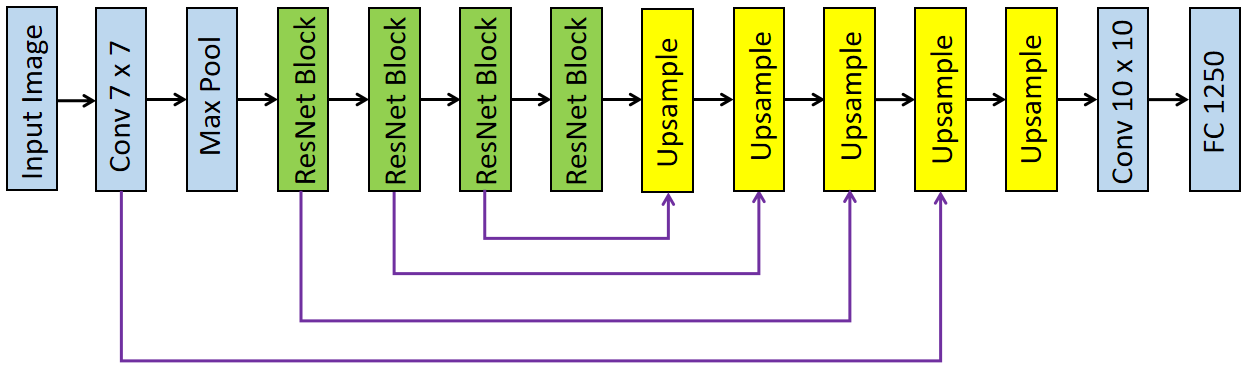}
    \caption{The architecture of ResNetUNet, designed by us for the displacement prediction of heterogeneous computational mechanics problems. The details of the architecture are provided in the supplementary section.}
    \label{fig:resnetunet}
\end{figure*}

The dataset is summarized in Tab. \ref{tab:2}.

\begin{table}[h]
\begin{center}
\caption{The dataset for the transient machine learning model.}
\label{tab:2}
\begin{tabular}{ c | c} 
\hline
 input image & ground truth/label \\
 \hline
  Input image with dip angle $0^\circ$ & 1000 time steps   \\ 
   & of numpy array (40 size) \\
 \hline
  $\vdots$$\quad$ (2998 examples) & $\vdots$$\quad$ (2998 examples)\\ 
 \hline
  Input image with dip angle $45^\circ$ & 1000 time steps   \\ 
   & of numpy array (40 size) \\
 \hline
\end{tabular}
\end{center}
\end{table}

To further prepare the data for the time series prediction models, we apply min-max scaling to the data:
\begin{equation}
    X =  \frac{X-X_{\min}}{X_{\max}-X_{\min}}.
\end{equation}

We also create the data sequences for the subsequent transformer and LSTM models by using five consecutive data points to predict the next data point, i.e.,
\begin{equation}
    u_{y,t},u_{y,t+1},u_{y,t+2},u_{y,t+3},u_{y,t+4}\Longrightarrow u_{y,t+5},
\end{equation}
where each $u_{y, t}$ represents a 1D array of size 40. We adopt sliding windows to generate the sequences, thus producing $89400$ displacement sequences. Similarly, we split the dataset into 95\% for training and 5\% for validation, and generate new samples (5\%) for testing.

\section{Methods}
In this project, we deal with two different computational problems: the static mechanics problem and the transient mechanics problem. We developed three models, including a carefully designed combination of the ResNet and UNet models to predict the static displacement, and the LSTM and transformer models that are based on the pretrained ResNetUNet model.

\subsection{Static mechanics model}
We are motivated to start from CNN models by the literature \cite{nie2020stress}, and we further test ResNet and UNet from a similar deep learning problem to handle the heterogeneity in the input images \cite{charng2020deep}. The CNN and ResNet networks reduce the dimension of the images layer by layer; they lose information while collecting important features from the input. Meanwhile, the deep neural network constructed with ResNet can be utilized to decode the physics of Eq. \ref{equ:stress}, which is a complicated procedure. Additionally, the heterogeneous distribution of the material affects the final result considerably. Inspired by segmentation in computer vision \cite{ronneberger2015u}, we use UNet to capture the heterogeneous distribution of the pixels/materials.

\subsubsection{Baseline models: CNN and ResNet}
As the baseline, we use the most basic CNN model, consisting of consecutive convolution blocks, each comprising a \texttt{conv2d}, a \texttt{BatchNorm2d}, and a \texttt{relu}. Finally, we flatten the convolutional output to obtain a fully connected layer with 1250 dimensions for the output. For the other baseline model, ResNet, we use the \texttt{resnet18} model in PyTorch and modify the last layer to have a convolution with width, height, and channel dimensions of $50\times 25 \times 1$, which is flattened into the fully connected layer afterwards.

\subsubsection{Model: ResNetUNet}
\begin{figure}[h]
    \centering
    \includegraphics[width = 0.6 \textwidth]{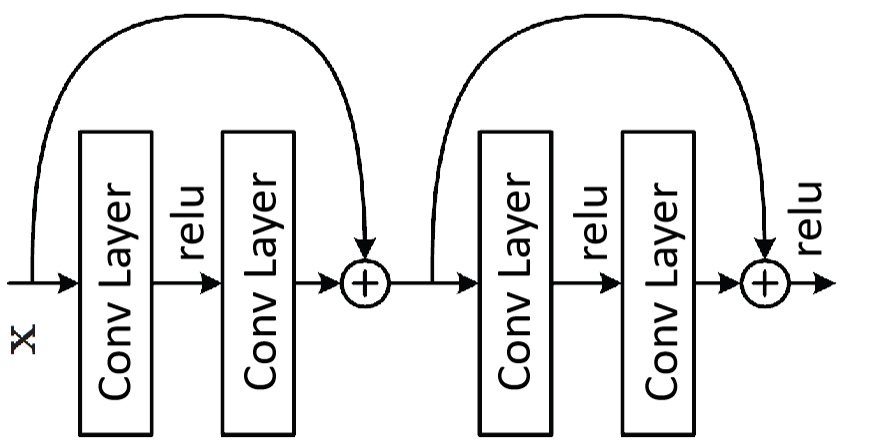}
    \caption{The implementation of the ResNet block in the ResNetUNet architecture.}
    \label{fig:rblock}
\end{figure}

ResNetUNet is a model designed by us to solve the current problem. Compared to the conventional UNet \cite{ronneberger2015u}, we modify the encoder part by using ResNet for downsampling the image to better capture the physical equations. The upsampling is the same as in the traditional UNet \cite{zhou2019d}. The architecture of ResNetUNet is illustrated in Fig. \ref{fig:resnetunet}. The special part here is the residual block, which replaces the ordinary CNN layers in UNet with two consecutive residual blocks, as shown in Fig. \ref{fig:rblock}. To successfully fulfill this functionality, we need to capture both the heterogeneous features and the physical laws. The UNet part performs segmentation-like work to distinguish the shale from the rock, while the ResNet part learns the complicated physical laws.

As for the loss function, we adopt the mean squared error (MSE) for all three models above:
\begin{equation}\label{equ:mse}
    \text{MSE}(y, \hat{y}) = \sum_{i=1}^{m}\frac{(y_i - \hat{y}_i)^2}{m},
\end{equation}
where $m$ is the number of examples. Note that we use the MSE as the loss function, but use both the MSE and the mean absolute error (MAE) as the evaluation metrics.

\begin{equation}\label{equ:mae}
    \text{MAE}(y, \hat{y}) = \sum_{i=1}^{m}\frac{|y_i - \hat{y}_i|}{m}.
\end{equation}

\begin{figure*}[h]
    \centering
    \includegraphics[width = 0.85 \textwidth]{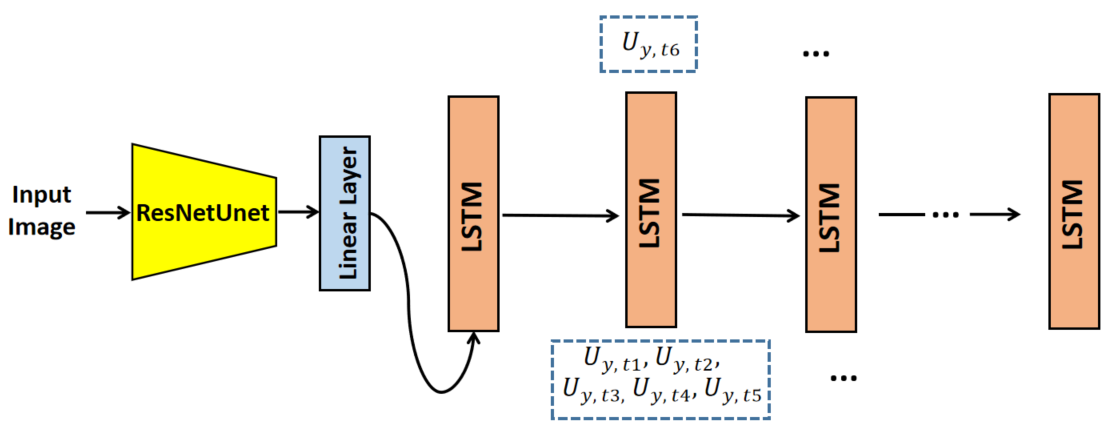}
    \caption{The architecture of the LSTM model to predict the evolution of the vertical displacement.}
    \label{fig:cnn-lstm}
\end{figure*}

\subsection{Transient mechanics model}
In this part, we use both the LSTM and the transformer to predict the evolution of the vertical displacement, the quantity of interest in industry, given the specified material distribution as an image. We adopt two models, LSTM and transformer, to perform the task, which is similar to image captioning \cite{tan2019phrase, liu2021cptr}, except that we use the displacement as the sequence rather than the word embeddings.

\subsubsection{LSTM model}
The combination of CNN and LSTM has been proven effective for image captioning \cite{tan2019phrase}. The CNN can be used to extract the features of the input image, which are fed into the LSTM network flow together with the sequence to be processed. In this project, we use a model pretrained on the static case to process the input geometry and obtain the features, which can embody the heterogeneous distribution of the material. The pretrained model used here is the ResNetUNet model trained on the 
static case, which has been proven effective for feature extraction. Then, we add a linear layer to enhance the expressivity of the model and to further process the features before they enter the LSTM network. The linear layer has the architecture of \texttt{in\_features=1250} and \texttt{out\_features=40}.

For the LSTM model, we use five steps of data points as the input to predict the next single step. Since the dimension of the sequence is 40, the number of points sampled on the ground surface, we set the input size of the sequence to 40. We adopt the \texttt{LSTMCell} model in PyTorch, which includes an input modulation gate, an input gate, a forget gate, and an output gate \cite{braun2018lstm}; it is initialized by setting the input size to 40 and the hidden size to 4. We stack four \texttt{LSTMCell}s to construct the model. The main architecture is depicted in Fig. \ref{fig:cnn-lstm}. The loss function used in this model is the MSE, the same as in Eq. \ref{equ:mse}. We also use both the MSE and the MAE as the metrics to evaluate the model. 

The architecture of this model is inspired by the LSTM image captioning model \cite{soh2016learning}. The difference lies in that we use the continuous values of the displacement as the data, rather than the \texttt{softmax} classification of the word embeddings. Thus, the loss functions for the two tasks are also distinct. Based on the accuracy requirements from industry, the dimensions of the sequence data in our engineering application are approximately the same as the dimensions of words in natural language processing. The similarity between the two models gives us the confidence to build our machine learning model based on the image captioning model in NLP.

\subsubsection{Transformer model}
For the transformer model, we also use the pretrained ResNetUNet to extract the image features, which are fed into a linear layer for further processing. The output of the linear layer then enters the transformer encoder. As for the decoder input, we use five consecutive sequences as the input and the next five consecutive sequences as the output. For example, we use $u_{y,t_1}, u_{y,t_2}, u_{y,t_3}, u_{y,t_4}, u_{y,t_5}$ as the input to predict the output $u_{y,t_2}, u_{y,t_3}, u_{y,t_4}, u_{y,t_5}, u_{y,t_6}$.

\begin{figure*}[h]
    \centering
    \includegraphics[width = 0.85 \textwidth]{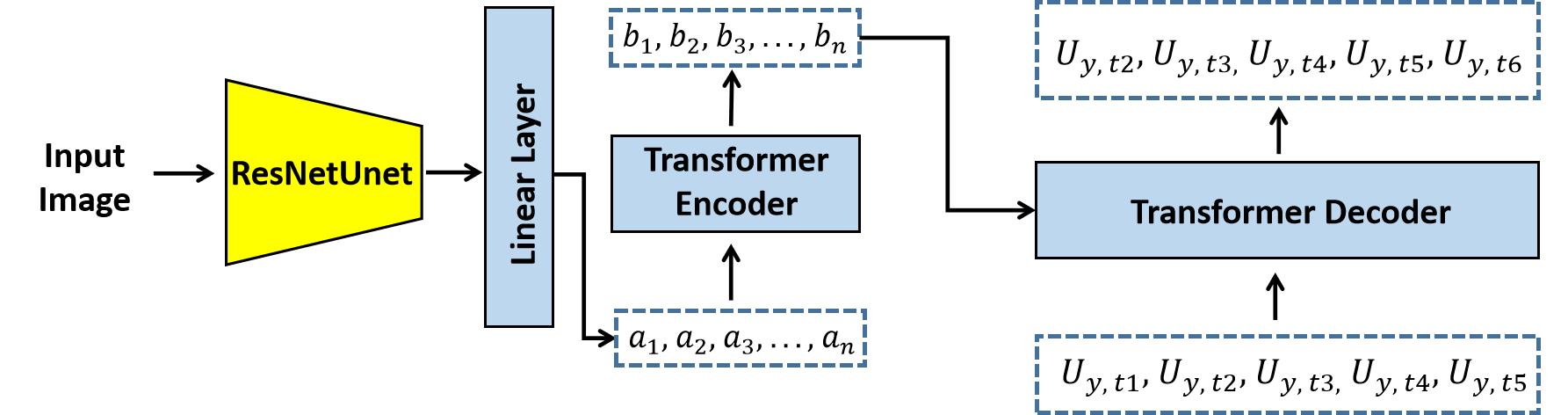}
    \caption{The architecture of the transformer model to predict the evolution of the vertical displacement.}
    \label{fig:cnn-transformer}
\end{figure*}

The input linear layer here has input features of size 1250 and output features of size 40, which reduces the dimension of the output of the pretrained model to 40. For the encoder part, we design another linear layer to convert the dimension from 40 to 16 to optimize memory usage. The output then enters the positional encoding layer to take the order of the sequence into account. Then, we employ the \texttt{nn.TransformerEncoderLayer} in PyTorch to assemble the encoder. The structure of the encoder follows the standard from the attention paper \cite{vaswani2017attention}. The parameters we set for the encoder layer are
\texttt{d\_model}=16 and \texttt{nhead}=8. We stack four \texttt{nn.TransformerEncoderLayer}s to form the encoder. For the decoder, we use the \texttt{nn.TransformerDecoderLayer}, which has the structure of \texttt{d\_model}=16 and \texttt{nhead}=8, and we stack four of them to set up the decoder. At the end of the transformer, we add a linear mapping to map the inner dimension of the decoder to the sequence dimension. The main architecture of the transformer is illustrated in Fig. \ref{fig:cnn-transformer}.

\section{Results}


We have studied baseline models to predict the displacement images based on the static mechanics model. The baseline models (CNN and ResNet) are compared with the ResNetUNet model for discussion. Next, we implemented the LSTM and the transformer as two different models for the transient mechanics model, because this is a problem related to videos. The deep learning model predictions are compared with the physics-based modeling results, which are treated as the ground truth in this project. Owing to the sufficient number of training samples, we do not include cross-validation in this work.

\subsection{Static mechanics model}

We selected the CNN as a baseline model because it collects important information from the input images as features by filtering through all pixels. We found that different subsurface geometries have a strong correlation with the displacement images, which motivated us to try the CNN first and examine its performance. We generated a total of 10,000 samples and then selected 95\% for training and 5\% for validation. The mini-batch size is 32 because this fits our CPU/GPU memory, and we prioritize the model convergence speed in terms of a smaller number of epochs. Although a smaller mini-batch size can be less efficient in terms of gradient computation \cite{bengio2012practical}, a large batch may degrade the quality of the model because it tends to converge to sharp minima \cite{keskar2016large}. We tuned the hyperparameters, including the learning rate, optimizer, beta1, beta2, and weight decay, to obtain a decreasing validation loss versus the number of epochs. We also compared the training loss with the validation loss to avoid underfitting or overfitting problems. As a result, the learning rate is tuned to 5e-6, and the weight decay is 1e-5. We finally use \texttt{Adam} as our optimizer because it requires less memory with a large number of parameters. More importantly, its hyperparameters have intuitive interpretations, and it requires less tuning with low computational cost \cite{kingma2014adam}. 

ResNet is the second model we investigated for the static mechanics prediction. The motivation for ResNet is high accuracy with moderate efficiency. It has a deeper neural network structure, so it is capable of expressing more complex models. We set the same mini-batch size and the same number of epochs for training, and we followed a similar procedure to determine the hyperparameters, including the learning rate, beta1, beta2, weight decay, and so on. The changes of the training loss and validation loss versus epochs help determine the hyperparameters.


\begin{figure}[h]
    \centering
    \includegraphics[width=0.9\textwidth]{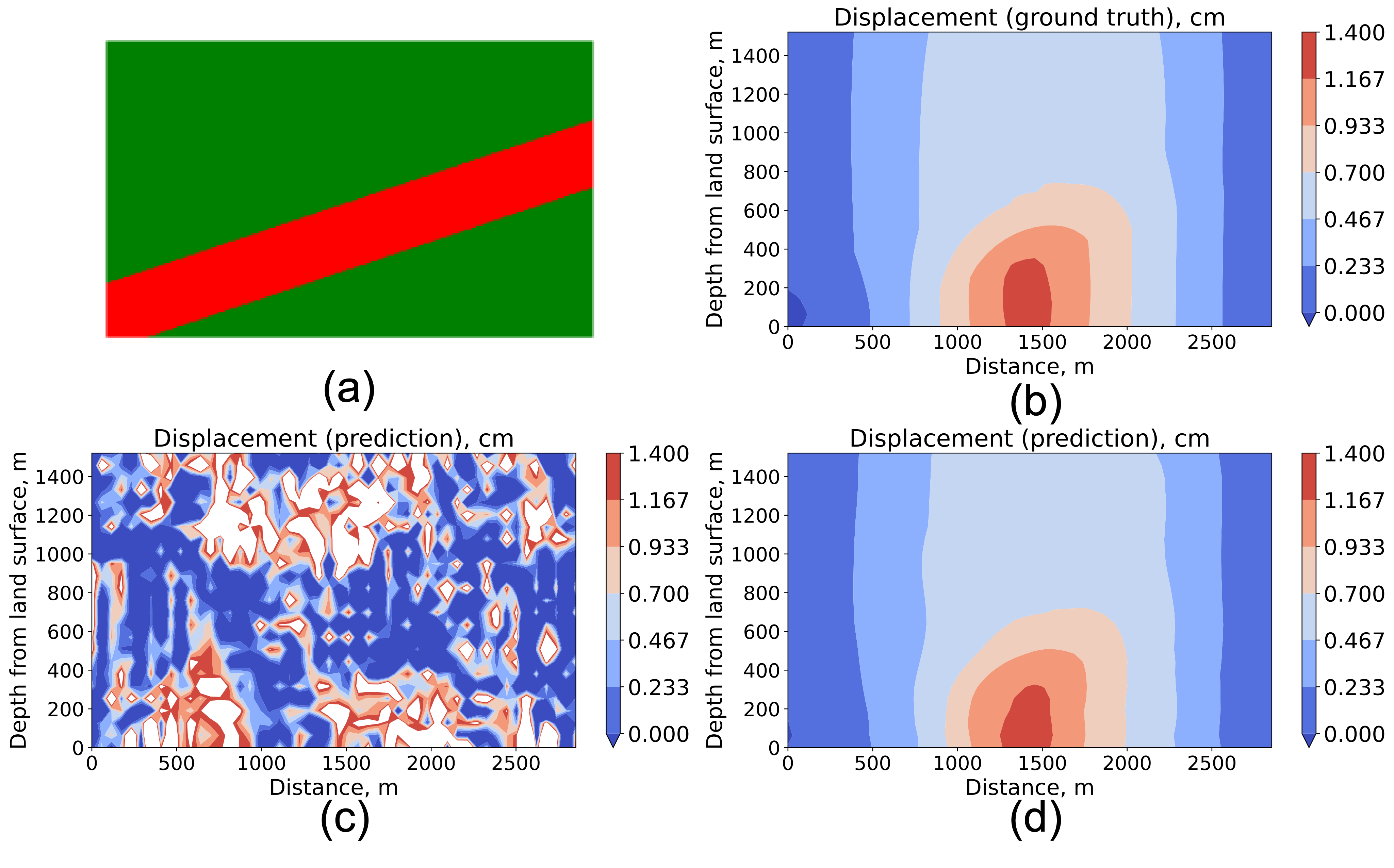}
    \caption{Comparison of the ResNet and ResNetUNet predictions of the displacement images based on the geometry image as input. (a) is the input image for the deep learning models, (b) is the ground truth based on physics, (c) is the prediction from our trained ResNet model, and (d) is the prediction from the ResNetUNet model.}
    \label{fig:resnet}
\end{figure}

The goal of this part is to develop a deep learning model that accurately predicts the displacement images. Therefore, it is important to visually compare the predictions with the ground truth side by side. Figure \ref{fig:resnet} shows the comparison of the performance of the different models. The color represents different values of the displacement as a physical quantity according to the color bar, and the unit is meters. From the ground truth image (Fig. \ref{fig:resnet} (b)), we observe that the largest displacement occurs close to the point where we apply the load, and the displacement becomes smaller farther away from the point of the applied stress. The reason the contour is not symmetric is that we consider heterogeneity in this system, meaning that the material of the subsurface rock is not the same at all locations. The motivation comes from subsurface engineers, who seek to characterize cases that better represent real-world scenarios. Our ResNet prediction (Fig. \ref{fig:resnet} (c)) illustrates the general trend of the displacement image on the right side of Fig. \ref{fig:resnet}. However, it clearly shows that the prediction is not sufficiently accurate. The results from the CNN model show comparable observations; therefore, we decided to use one failed case for analysis here. The reason is that ResNet and CNN downsample the input image to extract features, and the information loss through this process leads to the small regions without values in Fig. \ref{fig:resnet} (c).  

We are motivated to further improve the work by introducing the ResNetUNet model, because UNet includes an upsampling procedure that incorporates the information from each step of the downsampling process. This tends to avoid the information loss and provide more accurate predictions. We follow the same mini-batch design and again select Adam as the optimizer for this new model, and the same criterion is applied to determine the hyperparameters, including the learning rate, beta1, beta2, and weight decay. 

\begin{figure}[h]
    \centering
    \includegraphics[width=0.9 \textwidth]{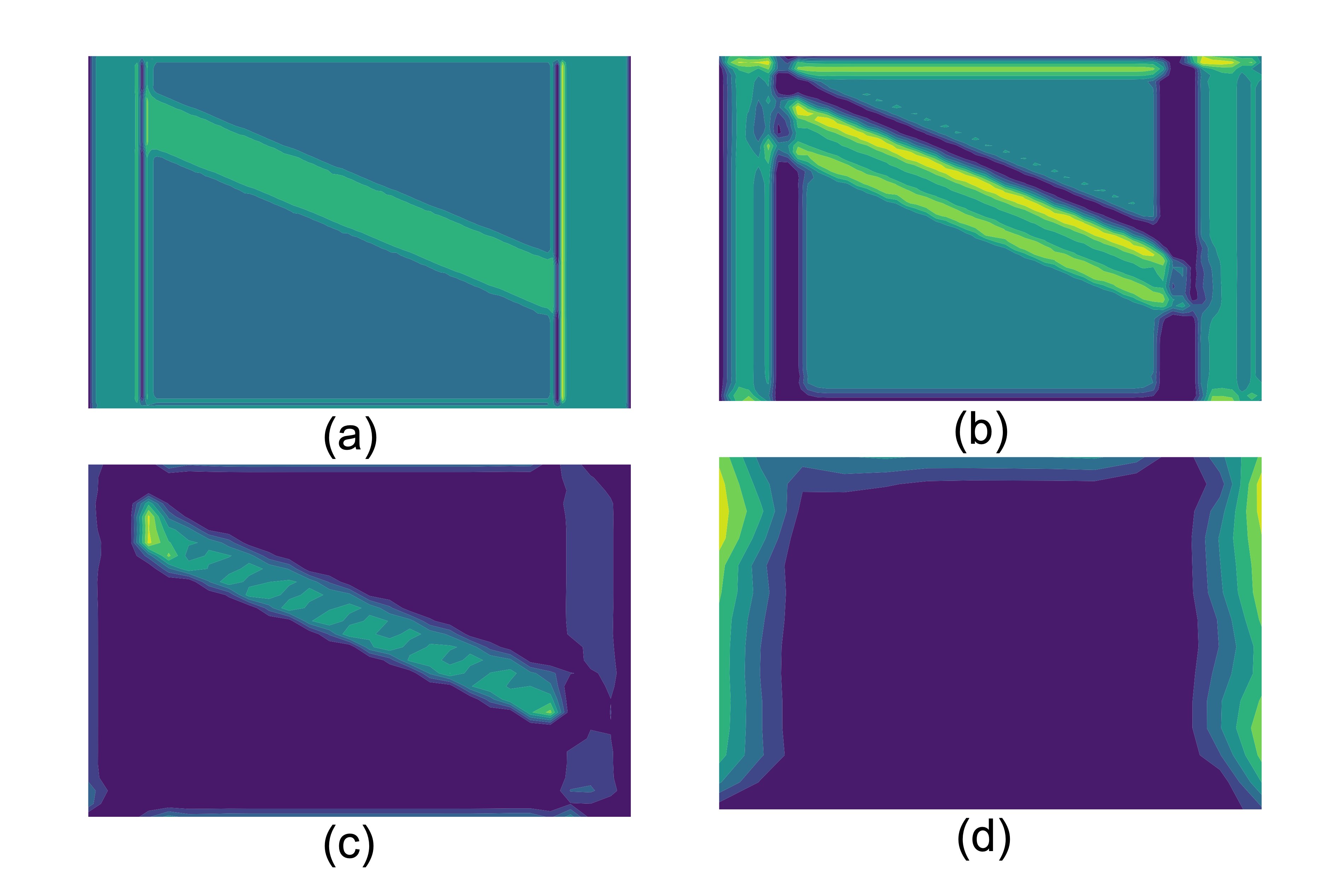}
    \caption{Intermediate outputs from the ResNetUNet decoding process.}
    \label{fig:UNetsteps}
\end{figure}

We tuned the learning rate to 2e-6, which proved sufficient, and we kept the other hyperparameters the same as for the CNN and ResNet models. Figure \ref{fig:resnet} (d) illustrates our results from the prediction based on the ResNetUNet model. We clearly observe a significant improvement in the predictions. The images are almost identical to each other, and the values are comparable, as represented by the colors of the contours in those images. The functionality of the ResNetUNet can be examined more deeply by extracting the intermediate features along the network flow, as depicted in Fig. \ref{fig:UNetsteps}. The four ResNet blocks are critical for capturing the features of the image. We can observe that the heterogeneity of the input geometry is captured first, and then the pixels begin to spread and oscillate, which is due to the physics that we need to learn. At the last ResNet block, all the information is compressed before being upsampled by the decoder part of the network.


\begin{figure}[h]
    \centering
    \includegraphics[width=0.9 \textwidth]{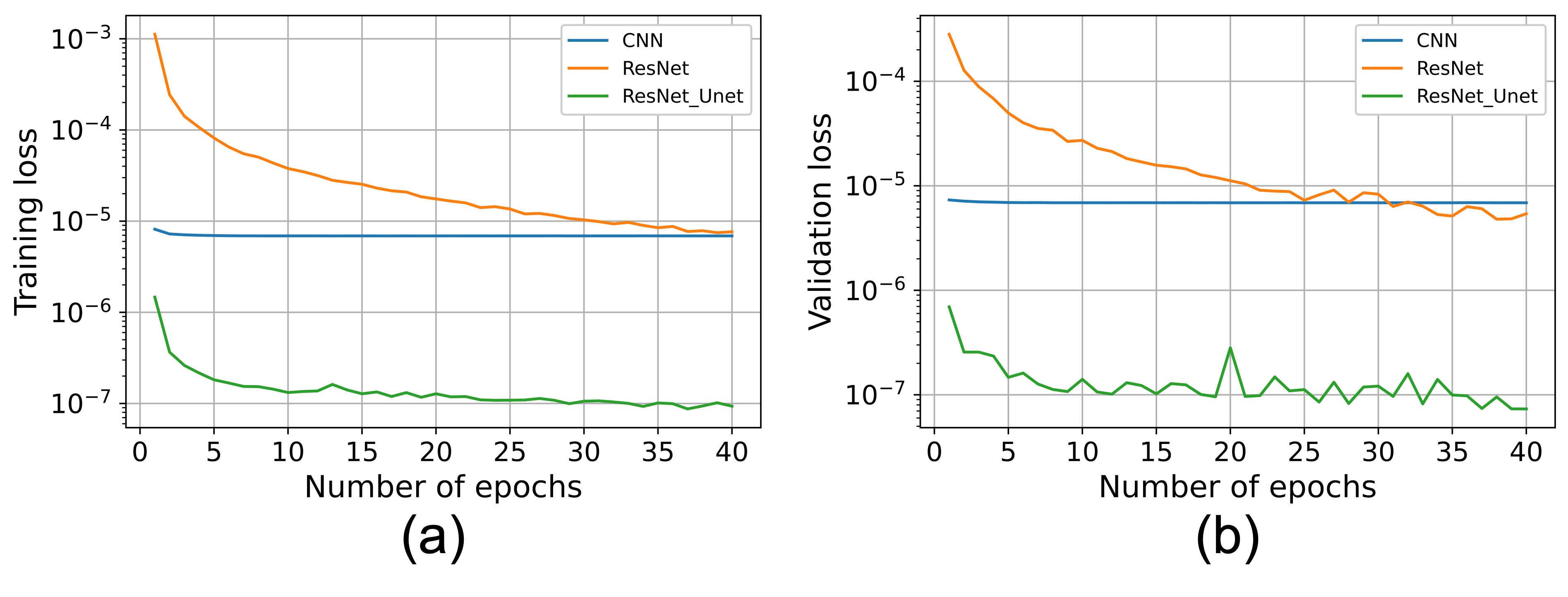}
    \caption{Training (a) and validation (b) loss versus epoch number for all investigated models, including CNN, ResNet, and ResNetUNet.}
    \label{fig:staticloss}
\end{figure}

The primary metrics to quantitatively evaluate the performance of the models are the MSE and the MAE. The ground truth comes from the physics-based simulation, and we are able to compare the displacement values at each point of the image between the prediction and the simulation. This effectively quantifies the accuracy of the model. In order to quantitatively evaluate the performance of the different models, we compare the training loss and the validation loss for all three models in Fig. \ref{fig:staticloss}. The loss history illustrates that our models are tuned appropriately, without overfitting or underfitting. The MSE and MAE errors are both collected and presented in Tab. \ref{tab:3} as well.


\begin{table}[h]
\begin{center}
\caption{Comparison of CNN, ResNet, and ResNetUNet using MSE and MAE loss values.}
\label{tab:3}
\begin{tabular}{ c|cc|cc }
\hline
\multirow{2}{*}{Model} & \multicolumn{2}{c|}{Training loss} & \multicolumn{2}{c}{Validation loss}   \\ \cline{2-5} 
                       & \multicolumn{1}{c|}{MSE}   & MAE   & \multicolumn{1}{c|}{MSE}     & MAE     \\ \hline
CNN                    & \multicolumn{1}{c|}{6.89e-6 }     &  2.67e-3   & \multicolumn{1}{c|}{6.88e-6} & 2.58e-3 \\ \hline
ResNet                 & \multicolumn{1}{c|}{7.85e-6}     & 2.22e-3     & \multicolumn{1}{c|}{4.79e-6} & 2.00e-3 \\ \hline
ResNetUNet             & \multicolumn{1}{c|}{7.82e-8}   & 1.90e-4    & \multicolumn{1}{c|}{7.33e-8} & 2.00e-4 \\ \hline
\end{tabular}
\end{center}
\end{table}

\subsection{Transient mechanics model}
It is essential to evaluate the transient mechanics model, because it is closer to real-world problems. The physics tells us that the displacement can change over time when a constant load is applied to a system of porous media with fluid inside. Therefore, the problem changes from a static image to a time series video problem. We are motivated to use the LSTM because it handles time series predictions well. Meanwhile, the transformer is becoming more and more influential, so we decided to compare the LSTM and transformer models. 

\begin{figure}[h]
    \centering
    \includegraphics[width=0.9 \textwidth]{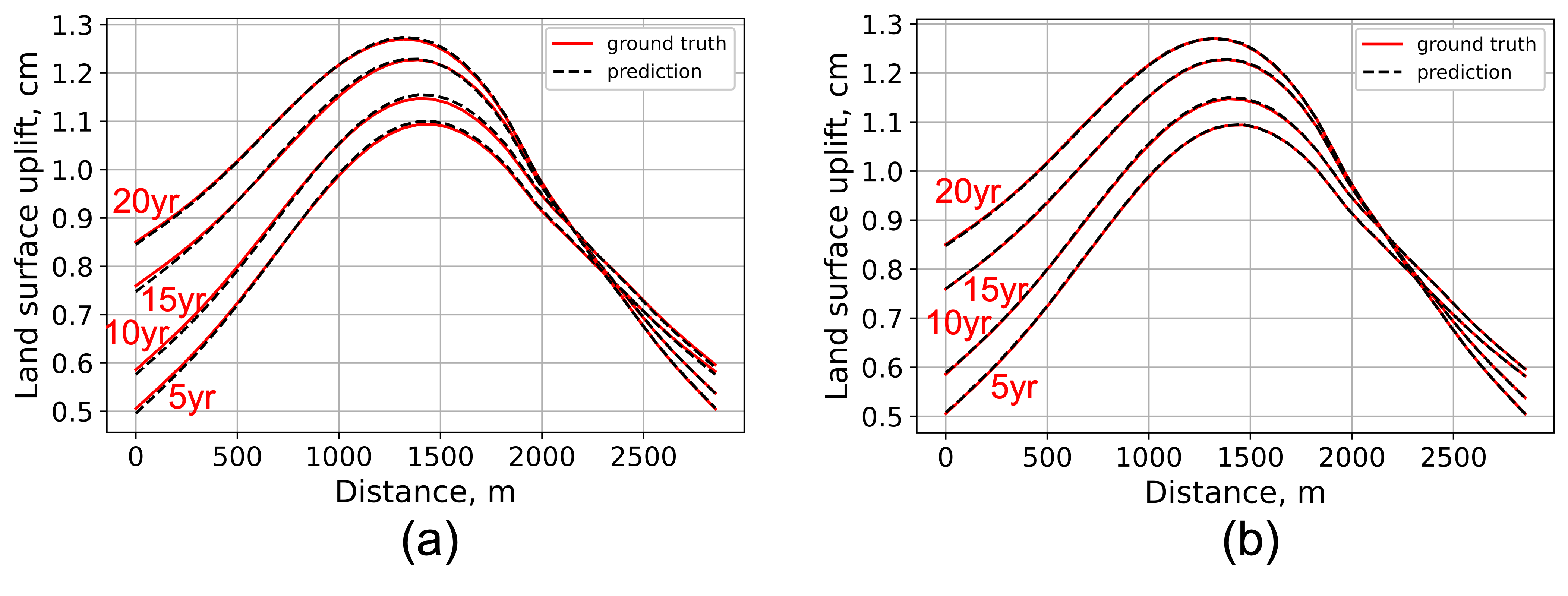}
    \caption{Predictions from (a) the LSTM and (b) the transformer for the transient problem. We compare the surface uplift across the site at 5 years, 10 years, 15 years, and 20 years after the load is applied in the subsurface.}
    \label{fig:transientpred}
\end{figure}

The LSTM model has shown promising potential for time series prediction in various fields \cite{sagheer2019time, cao2019financial}. However, the current data predictions mainly involve low-dimensional data, much lower than the case we deal with here, with a dimension of 40. For the training of the LSTM model, we tune parameters including the number of LSTM cells, the batch size, and the learning rate. Similarly, we use the \texttt{Adam} optimizer for gradient descent. To avoid overfitting and underfitting, we finally stack 4 LSTM cells, and we choose a batch size of 1024 to facilitate the execution on the GPU. The learning rate is chosen as 0.001. 

\begin{figure}[h]
    \centering
    \includegraphics[width=0.9 \textwidth]{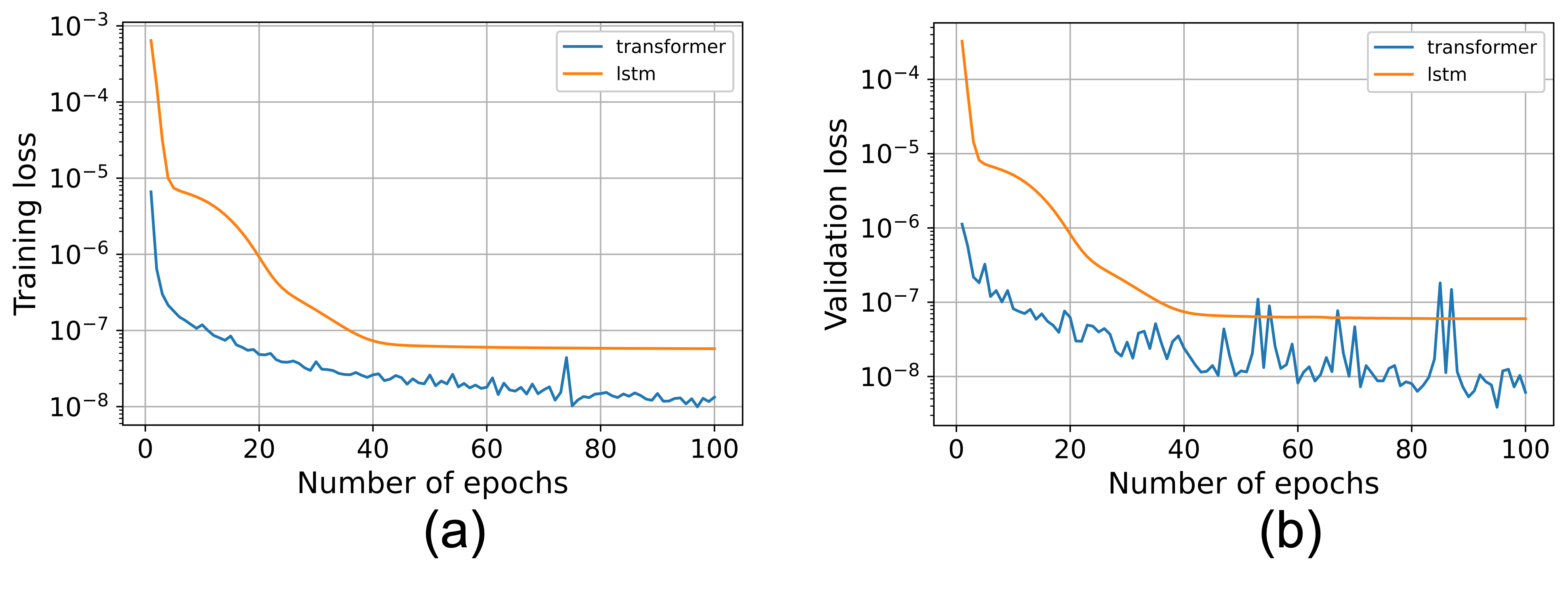}
    \caption{Comparison of model performance according to (a) training loss and (b) validation loss.}
    \label{fig:transientloss}
\end{figure}

The validation loss history is shown in Fig. \ref{fig:transientloss}. The model is explored extensively, and we found that the LSTM prediction can fit the ground truth with slight deviations (Fig. \ref{fig:transientpred}). This is because we are predicting a value-based output with a high accuracy requirement, and the high dimension of the sequence requires an appropriate allocation of attention from the model. For example, the displacements in the middle of the domain have larger variations than the displacements at the corners, thus requiring more learning. The attention mechanism is not capable of dealing with such a high-dimensional data prediction task. This also inspires us to utilize a model with more attention, which gives rise to the transformer. 

The transformer is carefully tuned with respect to the hyperparameters, such as the learning rate and the batch size, and the optimal values of these two parameters are the same as those of the LSTM. We also set the dropout probability inside the transformer model to 0.3. After running the same 100 epochs, the transformer performs better than the LSTM, with a lower loss, and its prediction results also fit the ground truth nearly perfectly. The loss history is illustrated in Fig. \ref{fig:transientloss}. To better quantify the LSTM and transformer models, we list the MSE and the MAE in Tab. \ref{tab:timeloss}, which shows that the transformer performs better than the LSTM in this case as well.

\begin{table}[h]
\begin{center}
\caption{Comparison of LSTM and transformer using MSE and MAE loss values.}
\label{tab:timeloss}
\begin{tabular}{ c|cc|cc }
\hline
\multirow{2}{*}{Model} & \multicolumn{2}{c|}{Training loss} & \multicolumn{2}{c}{Validation loss}   \\ \cline{2-5} 
                       & \multicolumn{1}{c|}{MSE}   & MAE   & \multicolumn{1}{c|}{MSE}     & MAE     \\ \hline
LSTM                    & \multicolumn{1}{c|}{6.89e-6 }     &  6.88e-6   & \multicolumn{1}{c|}{6.00e-8} & 6.11e-6 \\ \hline
Transformer                 & \multicolumn{1}{c|}{7.85e-6}     & 2.81e-6     & \multicolumn{1}{c|}{3.86e-9} & 1.50e-6 \\ 
\hline
\end{tabular}
\end{center}
\end{table}

\section{Conclusion and Future Work}

This paper demonstrates a state-of-the-art approach that leverages computer vision intelligence for geomechanics computations. We generalize our problem to arbitrary input geometry images, so our trained models are no longer restricted to a specific case or field. 

We observe that the ResNetUNet model stands out compared with the CNN and ResNet models for static displacement image prediction. The particular architecture of ResNetUNet ensures its excellent performance by passing the information from the encoding to the decoding process. In comparison, the CNN and ResNet models lose part of the information, although they successfully extract key features from the input. The LSTM and transformer models work well for the transient part of our work. We observe that training a model for video problems requires more RAM because more parameters are involved. For all the models, we used both the MSE and the MAE to quantify the loss, and the comparison of the training and validation losses indicates no underfitting or overfitting concerns. The LSTM model incurs lower computational costs during training, but at the price of slightly lower prediction accuracy. Both the LSTM and transformer models show strong potential for solving complex coupled solid mechanics and fluid flow problems, which is critical for carbon storage as one of the key components to mitigate global warming and protect our climate.

For future work, we will continue along this path and generate more complicated input geometry images using generative adversarial networks (GANs) to test the performance of our models. The goal is to make our models robust for land surface displacement prediction to inform decision making in carbon storage projects. In addition, we would like to launch an alpha version of a web application for users to access our models, so that our work will contribute value to society.

\section{Supplementary Material}

Our source code is available on GitHub (\href{https://github.com/AlexCHEN-Engineer/CNN_FEM}{click here}). 

For the time series prediction, refer to the demo (\href{https://drive.google.com/file/d/1-xafByddGVI42P0At4XCiMT_rSu0eUP1/view?usp=sharing}{click here}) to view the evolution of the displacement.

\section{Acknowledgements}

The authors thank Prof. Anthony Kovscek and Prof. Ronaldo Borja for valuable suggestions and advice in terms of carbon storage and rock mechanics. We acknowledge the insightful discussions with Ruohan Gao to improve our work on both image and video problems, and we also thank the support from Manasi, Sumith to formulate our work.

{\small
\bibliographystyle{ieee_fullname}
\bibliography{carbon_net_paper}
}

\section*{Appendix A. Physics formulation}
For the static model, the mathematical formulation is as follows. We solve a partial differential equation problem stated as: for the domain $\Omega$, the boundary of the domain is denoted by $\Gamma = \Gamma_u\cup \Gamma_t$, where $\Gamma_u\cap \Gamma_t=\emptyset$, and the partial differential equations we solve can be expressed as
\begin{equation}\label{equ:stress}
\nabla\cdot \boldsymbol{\sigma}+\rho\boldsymbol{g}=\boldsymbol{0},\ \boldsymbol{u} = \overline{\boldsymbol{u}}\ \text{on} \ \Gamma_u, \ \boldsymbol{\sigma}\cdot\boldsymbol{n} = \overline{\boldsymbol{t}}\ \text{on}\ \Gamma_t 
\end{equation}
where $\boldsymbol{u}$ refers to the displacement of the domain, $\boldsymbol{g}$ is the gravitational acceleration, and the stress $\boldsymbol\sigma$ above is expressed as
\begin{equation}
    \boldsymbol{\sigma} = \mathbb{C}^e:(\nabla\boldsymbol{u}+\boldsymbol{u}\nabla),
\end{equation}
where $\mathbb{C}^e$ is the stiffness matrix.

For the transient model, the mathematical formulation is:
\begin{subequations}
\begin{align}
&\nabla\cdot \boldsymbol{\sigma}+\rho\boldsymbol{g}=\boldsymbol{0},\ \boldsymbol{u} = \overline{\boldsymbol{u}}\ \text{on} \ \Gamma_u, \ \boldsymbol{\sigma}\cdot\boldsymbol{n} = \overline{\boldsymbol{t}}\ \text{on}\ \Gamma_t \\
&\nabla\cdot\dot{\boldsymbol{u}}+\nabla\cdot{\boldsymbol{q}}=0,\ p = \overline{p}\ \text{on} \ \Gamma_p,\ \boldsymbol{q} = \overline{\boldsymbol{q}}\ \text{on}\ \Gamma_q,
\end{align}
\end{subequations}
where $\Gamma_p$ and $\Gamma_q$ are the essential and natural boundaries for the pore pressure, and the stress $\boldsymbol{\sigma}$ is expressed as
\begin{equation}
    \begin{aligned}
    &\boldsymbol{\sigma} = \boldsymbol{\sigma}^\prime - p\mathbb{I},\\
    &\boldsymbol{\sigma}^\prime = \mathbb{C}^e:(\nabla \boldsymbol{u} + \boldsymbol{u}\nabla).
    \end{aligned}
\end{equation}

In addition to the symbols used in the static part, we have the pore pressure of the fluid defined over the domain as $p$, and the fluid flux $\boldsymbol{q}=-\boldsymbol{\kappa}/\mu \nabla p$, which is Darcy's law that associates the flux with the pore pressure through the permeability $\boldsymbol{\kappa}$ and the dynamic viscosity $\mu$.

\end{document}